\documentclass[10pt,twocolumn,letterpaper]{article}

\usepackage[pagenumbers]{cvpr}

\usepackage{capt-of}
\usepackage{xspace}
\usepackage{xcolor}
\usepackage{enumitem}
\usepackage{amsmath,amsthm,amssymb,amsfonts,dsfont,pifont,bm,bbm,mathrsfs,mathtools,nicefrac,extarrows,relsize}
\usepackage{algorithm,algpseudocode,listings}
\usepackage{booktabs,multirow,adjustbox,diagbox,threeparttable,tabularray,setspace}

\definecolor{cvprblue}{rgb}{0.21,0.49,0.74}
\usepackage[pagebackref,breaklinks,colorlinks,citecolor=cvprblue,bookmarks=false]{hyperref}
\usepackage{wrapfig}
\usepackage[capitalize]{cleveref}  

\crefname{section}{Sec.}{Secs.}
\Crefname{section}{Section}{Sections}
\crefname{appendix}{App.}{Apps.}
\Crefname{appendix}{Appendix}{Appendices}
\crefname{table}{Tab.}{Tabs.}
\Crefname{table}{Table}{Tables}
\crefname{figure}{Fig.}{Figs.}
\Crefname{figure}{Figure}{Figures}
\crefname{equation}{Eq.}{Eqs.}
\Crefname{equation}{Equation}{Equations}
\crefname{theorem}{Thm.}{Thms.}
\Crefname{theorem}{Theorem}{Theorems}
\crefname{lemma}{Lem.}{Lems.}
\Crefname{lemma}{Lemma}{Lemmas}
\crefname{remark}{Rem.}{Rems.}
\Crefname{remark}{Remark}{Remarks}
\crefname{corollary}{Cor.}{Cors.}
\Crefname{corollary}{Corollary}{Corollaries}
\crefname{algorithm}{Alg.}{Algs.}
\Crefname{algorithm}{Algorithm}{Algorithms}
\definecolor{cellred}{RGB}{213, 123, 101}
\definecolor{cellgreen}{RGB}{0, 205, 0}
\definecolor{cellblue}{RGB}{54, 125, 189}
\definecolor{codegreen}{rgb}{0,0.6,0}
\definecolor{codegray}{rgb}{0.5,0.5,0.5}
\definecolor{codepurple}{rgb}{0.58,0,0.82}
\definecolor{backcolour}{rgb}{1.0,1.0,1.0}
\lstdefinestyle{mystyle}{
    backgroundcolor=\color{backcolour},
    commentstyle=\color{codegreen},
    keywordstyle=\color{magenta},
    numberstyle=\tiny\color{codegray},
    stringstyle=\color{codepurple},
    basicstyle=\ttfamily\scriptsize,
    breakatwhitespace=false,
    breaklines=true,
    captionpos=b,
    keepspaces=true,
    numbers=left,
    numbersep=5pt,
    showspaces=false,
    showstringspaces=false,
    showtabs=false,
    tabsize=2
}
\newcommand{\methodname}{Edicho}
\newcommand{\method}{\texttt{\methodname}\xspace}
\newcommand{\supp}{\textit{Appendix}\xspace}

\newcommand\nonumfootnote[1]{%
\begingroup%
\renewcommand\thefootnote{}\footnote{\hspace{-3.7pt}#1}%
    \addtocounter{footnote}{-1}%
\endgroup%
}

\title{Edicho: Consistent Image Editing in the Wild}

\author{
    Qingyan Bai$^{1}$\quad
    Hao Ouyang$^{2}$\quad
    Yinghao Xu$^{3}$\quad
    Qiuyu Wang$^2$\quad \\
    Ceyuan Yang$^4$\quad
    Ka Leong Cheng$^1$\quad
    Yujun Shen$^{2\dagger}$\quad
    Qifeng Chen$^{1\dagger}$ \\
    \\
    $^{1}$HKUST \quad
    $^{2}$Ant Group \quad
    $^{3}$Stanford University \quad
    $^{4}$CUHK \quad
}

\begin{document}

\twocolumn[{
\maketitle
\vspace{-20pt}
\begin{center}
    \centering
    \captionsetup{type=figure}
    \includegraphics[width=\textwidth]{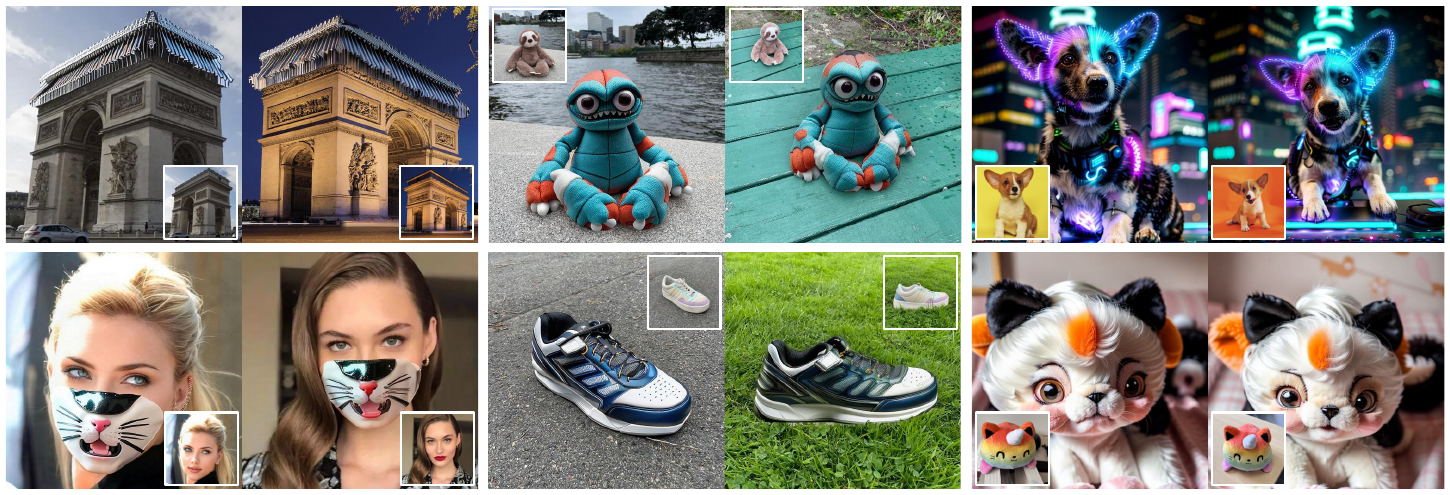}
    \vspace{-20pt}
    \captionof{figure}{Given two images in the wild, \method generates consistent editing versions of them in a zero-shot manner. 
    Our approach achieves precise consistency for editing parts (left), objects (middle), and the entire images (right) by leveraging explicit correspondence.
    }
    \label{fig:teaser}
    \vspace{2pt}
\end{center}
}
]
\nonumfootnote{$\dagger$ Corresponding authors.}

\begin{abstract}

As a verified need, consistent editing across in-the-wild images remains a technical challenge arising from various unmanageable factors, like object poses, lighting conditions, and photography environments.
\method\footnote{``\methodname'' is an abbreviation of ``edit echo'', implying that the edit is echoed across images.}
steps in with a training-free solution based on diffusion models, featuring a fundamental design principle of using \textbf{explicit} image correspondence to direct editing.
Specifically, the key components include an attention manipulation module and a carefully refined classifier-free guidance (CFG) denoising strategy, both of which take into account the pre-estimated correspondence.
Such an inference-time algorithm enjoys a plug-and-play nature and is compatible to most diffusion-based editing methods, such as ControlNet and BrushNet.
Extensive results demonstrate the efficacy of \method in consistent cross-image editing under diverse settings.
We will release the code to facilitate future studies.
Project page can be found \href{https://ant-research.github.io/edicho/}{here}.

\end{abstract}
\section{Introduction}
\label{sec:intro}

The ability to consistently edit images across different instances is of paramount importance in the field of computer vision and image processing~\cite{chen2024anydoor,alaluf2024cross, zhang2020cocosnet, zhou2021cocosnet2}. 
Consistent image editing facilitates numerous applications, such as creating coherent visual narratives and maintaining characteristics in marketing materials. 
As in~\cref{fig:teaser}, sellers or consumers can enhance photos of their favorite products, such as toys or shoes, by applying consistent decorative elements, making each item appear more appealing or personalized.
Similarly, during themed events like a masquerade ball or Halloween, families and friends may hope to uniformly style masks or dresses across their photos, ensuring a harmonious visual presentation.
Another instance for content creators is consistently making multiple of your photos look like a graceful elf or an impressive Superman.
By ensuring the edits applied to one image can be reliably replicated across others, we also enhance the efficiency and quality of tasks ranging from photo retouching to data augmentation for customization~\cite{ruiz2023dreambooth, kumari2023multi} and 3D reconstruction~\cite{wang2024dust3r}. 

Despite the significance of consistent editing, achieving it across diverse images remains a challenging task. 
Previous editing methods~\cite{zhang2023controlnet, ju2024brushnet, instructpix2pix} often operate on a per-image basis, leading to variations that can disrupt the uniformity required in specific applications. 
Previous attempts to address this issue have encountered limitations. 
Learning-based methods~\cite{yang2023paint, chen2024anydoor} that involve editing a single image and propagating the changes to others lack proper regularization and tend to produce inconsistent results.
They struggle to acquire high-quality paired training data and fail to enforce the necessary constraints to maintain uniformity. 
Alternatively, strategies without optimization~\cite{alaluf2024cross, cao2023masactrl, hertz2024stylealign} rely on the implicit correspondence predicted from the attention features to achieve appearance transfer.
Yet, due to the unstable implicit correspondence prediction, these approaches struggle to account for the intrinsic variations between images, leading to edits that appear inconsistent or distorted when applied indiscriminately. 

Inspired by the property of diffusion models~\cite{ho2020denoising, song2020ddim, rombach2022ldm, meng2021sdedit} where intermediate features are spatially aligned with the generated image space, we propose a novel, training-free, and plug-and-play method that enhances consistent image editing through explicit correspondence between images. 
Different from previous training-free methods~\cite{cao2023masactrl, hertz2024stylealign, alaluf2024cross} relying on implicit correspondence from attention weights to transfer appearance, we propose to predict the correspondence between the inputs with a robust correspondence extractor before editing.
Our approach then leverages the self-attention mechanism within neural networks to transfer features from a source image to a target image effectively.
Specifically, we enhance the self-attention mechanism by warping the query features according to the correspondence between the source and target images.
This allows us to borrow relevant attention features from the source image, ensuring that the edits remain coherent across different instances. 
To achieve finer control over the consistency of the edits, we further modify the classifier-free guidance (CFG)~\cite{dhariwal2021adm} computation by incorporating the pre-computed correspondence. 
This modification guides the generation process, aligning it more closely with the desired edits while maintaining high image quality. 
During this design, we empirically observed that directly transferring the source noisy latent to the target image often results in blurred and over-smoothed images. 
Inspired by the concept of NULL-text Inversion~\cite{mokady2023nulltext}, we discovered that fusing features from unconditional embeddings enhances consistency without compromising the image quality.

Moreover, our algorithm is specifically designed to handle in the wild images — those captured under diverse and uncontrolled real-world conditions. 
Benefiting from the correspondence, this capability ensures that our method remains robust against variations in lighting, backgrounds, perspectives, and occlusions commonly found in natural settings. 
By effectively processing in the wild images, 
the versatility of our method allows for additional numerous practical applications. 
For instance, in a customized generation, our method enables the generation of more consistent image sets by editing, which is valuable for learning customized models for novel concepts and creating personalized content.
Additionally, we can apply new textures consistently across different views of an object and acquire the corresponding 3D reconstructions of the edits, benefiting from the editing consistency.

In summary, we introduce explicit correspondence into the denoising process of diffusion models in order to achieve consistent image editing.
We enhance the self-attention mechanism and modify classifier-free guidance to incorporate correspondence information, improving edit consistency without degrading image quality.
We also further demonstrate that fusing features from unconditional embeddings enhances consistency, inspired by null-text inversion techniques.
The final method, due to its training-free and plug-and-play nature, is able to function across various models and diverse tasks, enabling both global and local edits.
We validate the effectiveness of the proposed method through extensive experiments, showing superior performance in both quantitative metrics and qualitative assessments.
\section{Related Work}
\label{sec:related_works}

\noindent\textbf{Generative models for image editing.}
Recently, diffusion models have shown unprecedented power in various generative tasks~\cite{nichol2021improved, song2020ddim, dhariwal2021adm, rombach2022ldm, karras2022edm, khachatryan2023text2video, chen2023videocrafter1, poole2022dreamfusion, ruiz2023dreambooth, hollein2023text2room, kumari2023multi, ho2022video, avrahami2023blended, blattmann2023svd, blattmann2023align, esser2023structure, wang2023rodin, kawar2023imagic, bai2024real, tewel2024consistory, voleti2024sv3d, cao2024instruction, cheng2023learning, ouyang2023codef}. 
To unleash its potential in editing, PnP~\cite{tumanyan2023pnp} proposes to borrow convolutional and attention features from the input image during generation to achieve manipulation.
While MasaCtrl~\cite{cao2023masactrl} and Cross-Imgae-Attention~\cite{alaluf2024cross} modify self-attention modules for editing, by combining the target queries and source keys and values.
Prompt2Prompt~\cite{mokady2022prompt2prompt} focuses on the cross-attention layers in text-to-image models and proposes manipulating the textual embedding.
Serving as the foundation of these editing methods, image inversion is also widely studied by researchers~\cite{song2020ddim, mokady2023nulltext, ju2023direct, miyake2023negative, han2024proxedit}.
Different from the aforementioned training-free editing methods, 
Instruct-Pix2Pix~\cite{instructpix2pix}, ControlNet~\cite{zhang2023controlnet, zhao2024uni}, T2I-Adapter~\cite{mou2024t2iadapter}, Composer~\cite{huang2023composer}, and BrushNet~\cite{ju2024brushnet} learn editing models conditioned on the input images and instructions, which are based on or fine-tuned from the pre-trained latent diffusion models for better quality and training stability.
Another branch of works~\cite{yang2023paint, chen2024anydoor, chen2024mimicbrush} aims at exemplar-based editing, where a pre-trained diffusion model is finetuned to function conditioned on the exemplar image as well as the masked source image.
\cite{nguyen2024edit} achieves pose transfer among the image batch by manipulating the StyleGAN latent codes following the exemplar image.
Unlike the works discussed above, we focus on the task of consistent editing for images in the wild, and propose an explicit correspondence-guided solution.

\noindent\textbf{Correspondence from neural networks.}
The concept of correspondence is widely applicable and essential in various real-world scenarios~\cite{zhang2020cocosnet, zhou2021cocosnet2}, where understanding relationships between data points is crucial.
Neural networks have been broadly employed to find correspondences for image, video, and 3D scenes through supervised learning~\cite{ranjan2017optical, jiang2021cotr, wang2024dust3r, leroy2024mast3r, jiang2021cotr}.
DIFT~\cite{tang2023dift} proposes to extract semantic correspondence among in-the-wild images by directly matching the features from the pre-trained diffusion models.
SD-DINO~\cite{zhang2024tale} further ensembles features from diffusion models and DINO~\cite{caron2021dino} for correspondence matching.
Once correspondences are established, they can be utilized in various applications. 
For instance, in object tracking, networks can maintain correspondences across video frames to follow objects through occlusions and transformations~\cite{doersch2023tapir, ouyang2023codef, karaev2023cotracker, xiao2024spatialtracker, sam-pt}. 
Our work leverages these principles by integrating correspondence into the diffusion model framework, enabling precise and consistent multi-image editing without additional training. 
\section{Method}
\label{sec:method}

\begin{figure}[t]
    \centering
    \includegraphics[width=1.0\linewidth]{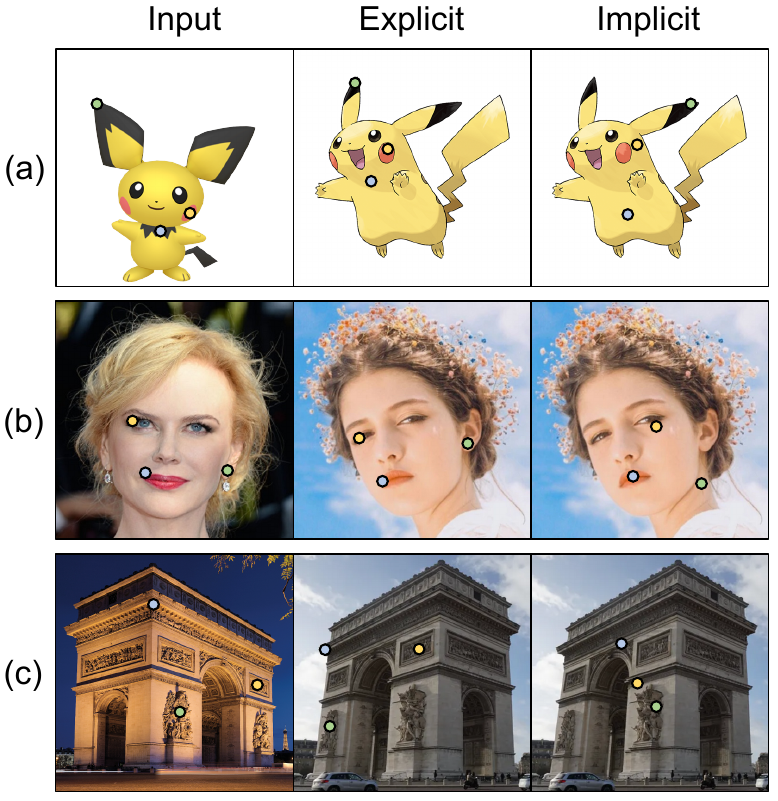}
    \caption{
      \textbf{Comparisons} of the implicit and our explicit correspondence prediction for the images in the wild.
      The implicit correspondence from cross-image attention calculation is less accurate and unstable with the change of denoising steps and network layers.
    } 
    \label{fig:corr_vis}
    \vspace{-4pt}
\end{figure}

\begin{figure*}[h]
    \centering
    \includegraphics[width=1.0\textwidth]{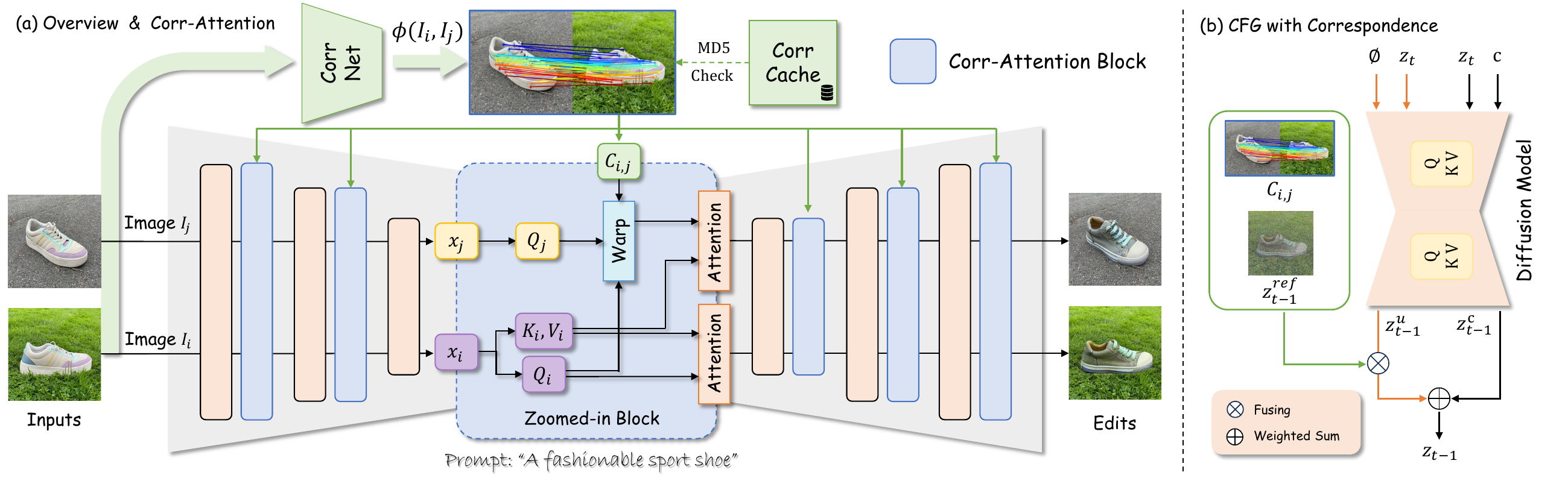}
    \caption{%
        \textbf{Framework} of~\method.
        To achieve consistent editing, we first predict the explicit correspondence with extractors for the input images.
        The pre-computed correspondence is injected into the pre-trained diffusion models and guide the denoising in the two levels of (a) attention features and (b) noisy latents in CFG.
    }
    \label{fig:framework}
    
    \vspace{-4pt}
\end{figure*}

In this work, we focus on the task of  consistent image editing, where multiple images are manipulated altogether to achieve consistent and unified looks.
To achieve this, we first extract explicit semantic correspondence among the image pairs by existing visual understanding methods such as~\cite{wang2024dust3r, tang2023dift, zhang2024tale, leroy2024mast3r}.
Then we seek help from the pre-trained editing models~\cite{zhang2023controlnet, ju2024brushnet} built upon Stable Diffusion~\cite{rombach2022ldm} to achieve editing, 
and guide their denoising process with these pre-computed explicit correspondences to ensure consistency.
In this section, we first review some preliminary concepts of diffusion models, which is followed by a subsection discussing the correspondence extraction and comparisons.
Then we introduce the correspondence-guided denoising process that includes two levels - the level of attention features and the level of noisy latents.
Note that these designs on feature manipulations are only applied to a range of denoising steps and layers, in order to preserve the strong generative prior from the pre-trained models.

\subsection{Preliminaries}
\noindent\textbf{Diffusion models} are probabilistic generative models trained through a process of progressively adding and then removing noise.
The forwarding process adds noise to the images as follows:
\begin{equation}
\boldsymbol{x}_t=\sqrt{\alpha_t} \cdot \boldsymbol{x}_0+\sqrt{1-\alpha_t} \cdot \boldsymbol{z},
\end{equation}
where $\boldsymbol{z}\sim\mathcal{N}(0, \mathbf{I})$ and $\alpha_t$ indicates the noise schedule.
And a neural network $\boldsymbol{\epsilon}_\theta\left(\boldsymbol{x}_t, t\right)$ is trained to predict the adding noise $\boldsymbol{z}$ during the denoising backward process and finally achieves sampling from Gaussian noise $\boldsymbol{x}_T\sim\mathcal{N}(0, \mathbf{I})$.
In the formulation of latent diffusion models (LDMs)~\cite{rombach2022ldm}, a pair of pre-trained variational encoder $\mathcal{E}$ and decoder $\mathcal{D}$ serve perceptual compression and enable denoising from the noisy latents $z$ in this latent space.

\noindent\textbf{Classifier-free guidance (CFG)}~\cite{dhariwal2021adm} is an innovative technique introduced to enhance the quality and diversity of generated images by diffusion models without relying on additional classifiers. 
Specifically, CFG introduces a mixing coefficient to blend the conditional and unconditional predictions made by the denoising model. The unconditional prediction is typically obtained by setting the condition to a null or default value. 

\noindent\textbf{Reference networks for editing.} Recent editing methods~\cite{zhang2023controlnet, ju2024brushnet} purpose editing by learning an additional reference network over the pre-trained large diffusion models while keeping the pre-trained backbone fixed.
This network-topology-preserving design successfully separates control signals and the pretrained generative prior.

\subsection{Correspondence Comparison and Prediction}

\noindent\textbf{Correspondence comparisons.}
In order to achieve the target of consistent editing, we start with the comparison between explicit and implicit correspondence for matching.
Explicit extractors predict correspondence from the input images with a single-pass forwarding and apply this prediction to all the target network layers and denoising steps.
While implicit extractors predict correspondences by calculating the similarities between attention queries and keys for each layer and denoising step.
As in the previous training-free editing methods~\cite{alaluf2024cross}, 
these correspondences are subsequently applied to this layer and step for editing.

In~\cref{fig:corr_vis}, we present the correspondence prediction results using explicit and implicit methods. 
For explicit prediction in cases (a) and (b), we employed DIFT~\cite{tang2023dift} and Dust3R~\cite{wang2024dust3r} is utilized for (c). 
For the implicit approach, we followed Cross-Image-Attention~\cite{alaluf2024cross} to compute the correspondence based on the attention similarity by querying the attention keys of the matching image with $Q_{i}\cdot K_{j}^T$ and visualize the corresponding positions with maximum similarity, where $i$ and $j$ indicate the image indices.
Additionally, for cases (a), (b), and (c), we selected different network layers and denoising steps (1, 10), (2, 15), (4, 25) for extraction to achieve a more comprehensive exploration where x and y in (x, y) represent the decoder layer number of the diffusion model and the denoising step. 
The visualized results in~\cref{fig:corr_vis} indicate that the correspondences obtained by explicit prediction are notably more accurate than those obtained by implicit methods. 
Moreover, the predictions from implicit methods tend to become unstable with changes in network layers and denoising steps.
These results are also aligned with previous work~\cite{zhang2024tale, tang2023dift} indicating that only specific layers or steps of generative models are suitable for effective visual understanding like point matching.
The inaccurate correspondence matching leads to borrowing inaccurate features in performing cross-image attention, which hinders the editing consistency of the editing methods merely based on implicit attentions~\cite{cao2023masactrl, alaluf2024cross, hertz2024stylealign}.  
This further boosts our motivation of introducing more robust explicit correspondence to guide the denoising process.
Additional details and comparisons for correspondence prediction are provided in~\supp.

\noindent\textbf{Correspondence prediction.} 
To achieve consistent editing for images $I_{i}$ and $I_{j}$,
the first step in our editing method involves extracting robust correspondence from the input images using a pre-trained correspondence extractor such as~\cite{wang2024dust3r, tang2023dift}:
\begin{align}
    \mathcal{C}_{i, j} = \phi (I_{i}, I_{j}),
\end{align}
where $\phi$ and $\mathcal{C}$ indicate the extractor and correspondence.
In our practice, the extractor is instantiated as in DIFT~\cite{tang2023dift}.
To further optimize efficiency, we implement a strategy to avoid redundant computations of correspondences, particularly when the same images or image groups are processed multiple times. 
We achieve this by encoding each image group using an MD5~\cite{rivest1992md5} hash function, creating a unique identifier for each.
After storing the identifier (key) and correspondence (value) in a minor database, the input image group would first retrieve 
it before editing for acceleration.

\subsection{Attention Manipulation with Correspondence}
\label{subsec:corr_attention}
Recall that the intermediate feature $\rm\bf x_{i}$ in self-attention blocks are firstly projected to queries $Q_{i}=f_Q(\rm\bf x_{i})$, keys $K_{i}=f_K(\rm\bf x_{i})$, and values $V_{i}=f_V(\rm\bf x_{i})$ with the learned projection matrix $f_Q$, $f_K$, and $f_V$.
Then the attention features $F$ could be computed by autonomously computing and assessing the relevance of these feature representations following~\cite{vaswani2017attention}.
Inspired by the comparisons between explicit and implicit correspondence,
we propose to guide self-attention with explicit correspondence to achieve consistent editing, which is termed as Corr-Attention.
For an image pair $(I_{i}, I_{j})$ among the inputs, we borrow features from the query matrix $Q_{i}$ to $Q_{j}$ to form a new query $Q_{edit}$ based on the explicit correspondence:
\begin{align}
\label{eq:attention_warp1}
    Q_{edit} & = \textit{Warp}(Q_{i}, Q_{j}, \mathcal{C}_{i, j}), 
\end{align}
Where the $\textit{Warp}$ function indicates the process of borrowing features by warping corresponding tokens to the source based on the corresponding location denoted by correspondence. 
Considering (1) tokens of $Q_{edit}$ are borrowed from $Q_{i}$ and (2) to further improve consistency, we query $K_{i}, V_{i}$ instead of $K_{j}, V_{j}$ during the editing of $I_{j}$:
\begin{align}
\label{eq:attention_warp2}
    F_{j} & = \text{softmax} \left ( \frac{Q_{edit}\cdot K_{i}^T}{\sqrt{d_{k}}} \right ) \cdot V_{i},
\end{align}
where $d_{k}$ indicates the dimension of $Q$ and $K$, and $F_{j}$ represents the attention outputs of $I_{j}$.
By transferring attention features from the source, we effectively achieve editing consistency during the denoising process.

\subsection{Classifer-free Guidance with Correspondence}
\label{subsec:corr_CFG}
In order to retain a finer consistency over the edited images, we take a further step from the attention feature control and focus on the noisy latents in Classifier-free Guidance (CFG)~\cite{dhariwal2021adm}.
Specifically, we extend the traditional CFG framework to facilitate synchronized editing of multiple images by leveraging explicit correspondences and propose Corr-CFG.
NULL-text inversion~\cite{mokady2023nulltext} demonstrates that optimizing unconditional word embeddings can achieve precise image inversion and semantic editing.
Inspired by this approach, our primary objective is to preserve the integrity of the pre-trained model's powerful generative priors during the consistent editing process.
To achieve this, we propose manipulating only the unconditional branch of $z_{j}$ within the Classifier-Free Guidance (CFG) framework under the guidance of correspondence.
Recall that in CFG the denoising process are split into two branches of the conditional and unconditional, and the noise are estimated with neural network $\epsilon_{\theta}$:
\begin{align}
\mathbf{z}^{c}_{t-1} &= \epsilon_{\theta}\left(\mathbf{z}_{t}, c\right),\\
    \mathbf{z}^{u}_{t-1} &= \epsilon_{\theta}\left(\mathbf{z}_{t}, \varnothing\right),
\end{align}
where $c$ represents the condition (text prompt) and $\varnothing$ indicates the null text.
Specifically, we modified the unconditional noise component of $z_{j}$ and incorporated information from $z_{i}$ into it during the denoising process, which ensures coherent edits:
\begin{align}
    \epsilon_{\theta}^{\text {uncond }}\left(\mathbf{z}_{t}^{j}\right)=\mathcal{T}\left(\epsilon_{\theta}\left(\mathbf{z}_{t}^{i}, \varnothing\right), 
    \epsilon_{\theta}\left(\mathbf{z}_{t}^{j}, \varnothing\right),
    \mathcal{C}_{i, j}\right),
\end{align}
where $\mathcal{T}$ represents a fusing function that aligns the unconditional noises and $t$ indicates the time-step:
\begin{align}
    \mathcal{T}(z_{i}, z_{j}, \mathcal{C}_{i, j}) = (1-\lambda) \cdot z_{j} + \lambda \cdot Inj(z_{i}, z_{j}, \mathcal{C}_{i, j}, \gamma).
\end{align}
$\lambda$ and $\gamma \in (0, 1]$ here are adjustable parameters. The function $Inj$ indicates randomly choosing latent of $z_{i}$ in a portion of $\gamma$ and injecting them into  $z_{j}$.
At last, we apply the guidance and fuse the conditional and unconditional predictions as in the prior paradigm~\cite{dhariwal2021adm}:
\begin{equation}
        \epsilon_{\theta}^{\text {guided }}\left(\mathbf{z}_{t}, c\right)=\epsilon_{\theta}^{\text {uncond }}\left(\mathbf{z}_{t}\right) + \\ s \cdot\left(\epsilon_{\theta}\left(\mathbf{z}_{t}, c\right)-\epsilon_{\theta}\left(\mathbf{z}_{t}, \varnothing\right)\right),
\end{equation}
where $s$ indicates the guidance scale.
The final latents generated as such are at last sent to the VAE decoder~\cite{rombach2022ldm} to be decoded into images.
\begin{figure*}[h]
    \centering
    \includegraphics[width=1.0\textwidth]{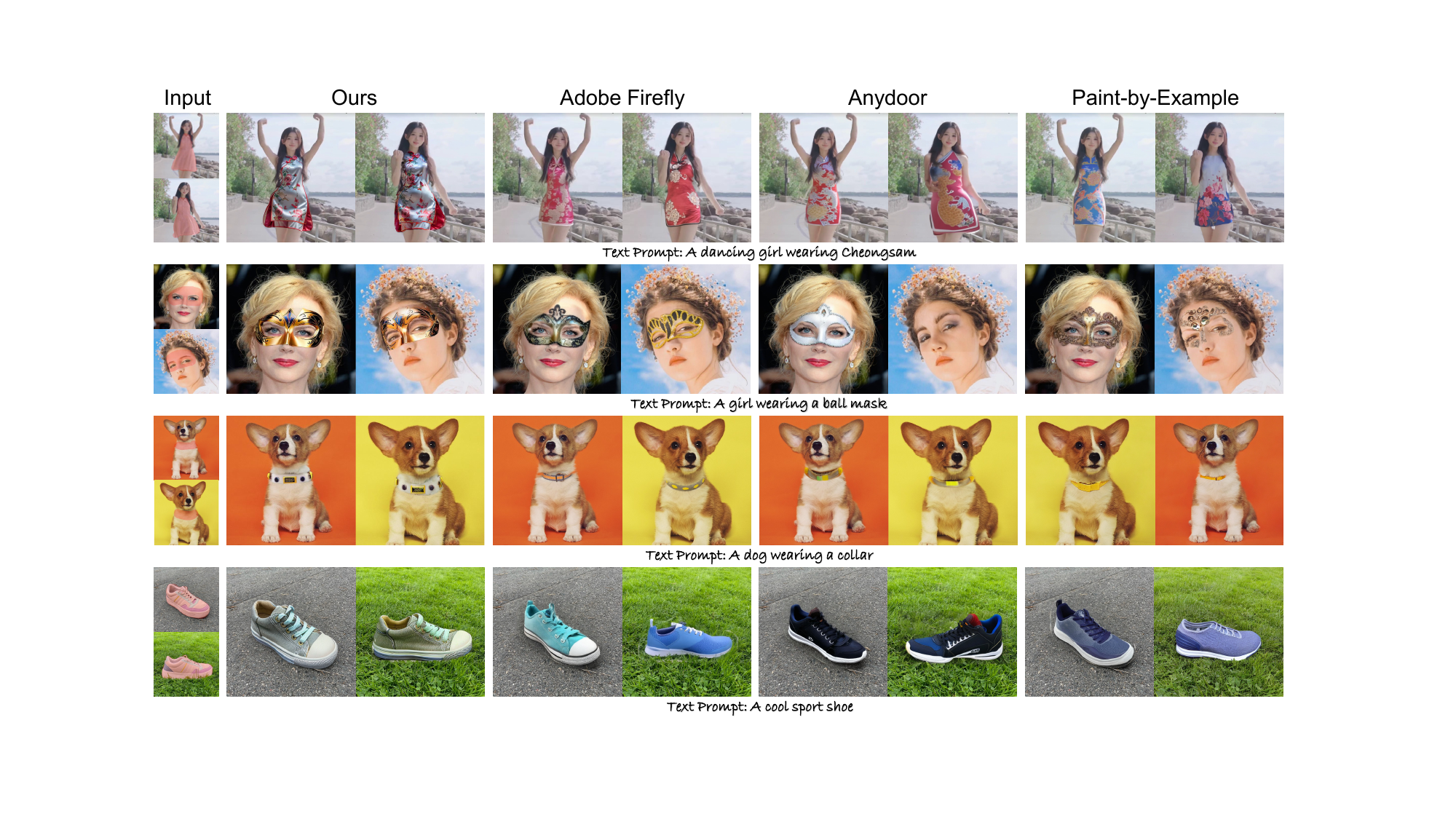}
    \caption{
    Qualitative comparisons on local editing with Adobe Firefly (AF)~\cite{Firefly}, Anydoor (AD)~\cite{chen2024anydoor}, and Paint-by-Example (PBE)~\cite{yang2023paint}.
    The inpainted areas of the inputs are highlighted in red.
    }
    \label{fig:local_comp}
\end{figure*}

\begin{figure*}[h]
    \centering
    \includegraphics[width=1.0\textwidth]{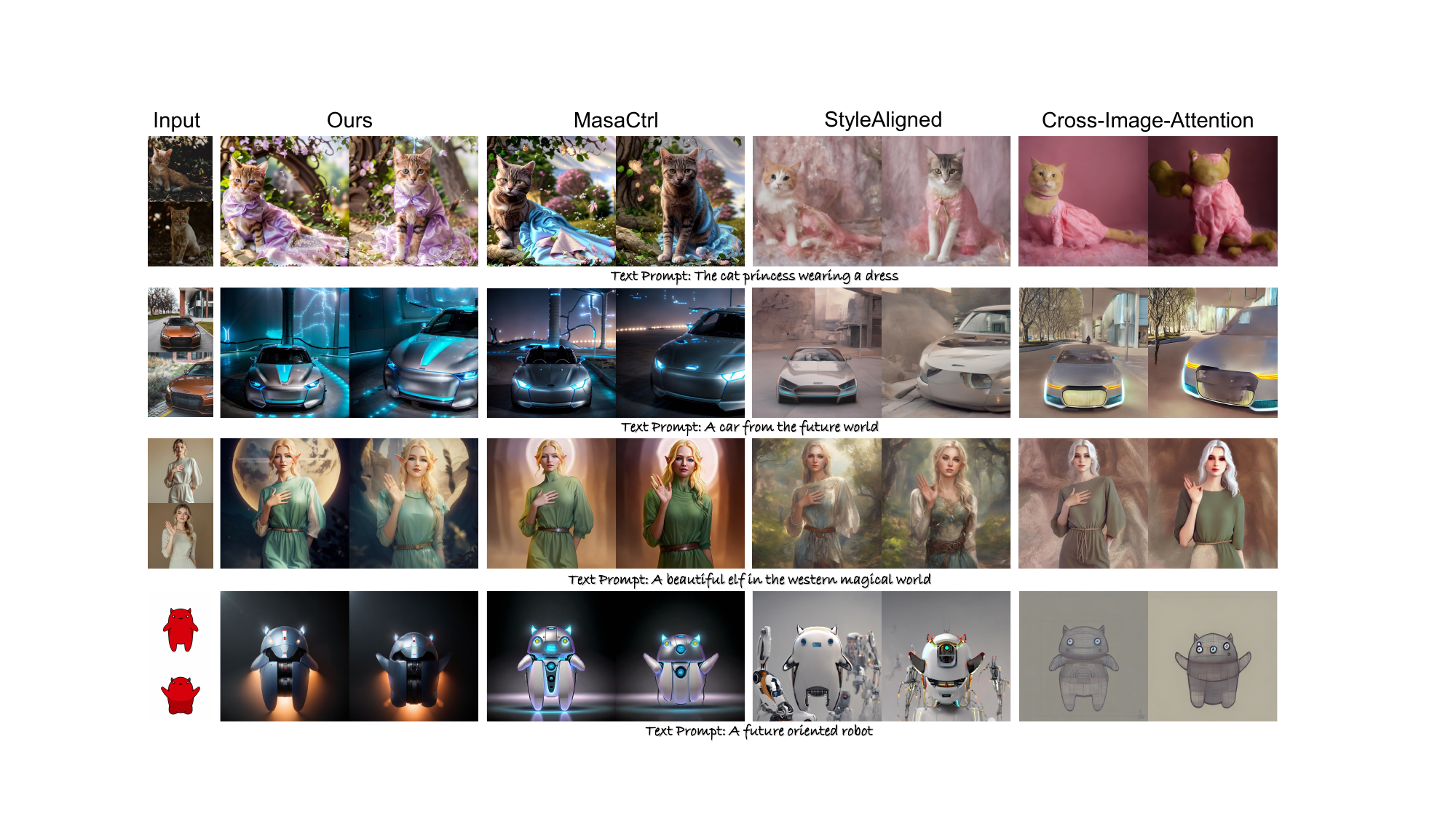}
    \caption{
    Qualitative comparisons on global editing with MasaCtrl (MC)~\cite{cao2023masactrl}, StyleAligned (SA)~\cite{hertz2024stylealign}, and Cross-Image-Attention (CIA)~\cite{alaluf2024cross}.
    }
    \label{fig:global_comp}
\end{figure*}

\section{Experiments}
\label{sec:exp}

\subsection{Experimental Setup}
\label{subsec:settings}
\noindent\textbf{Settings}. 
We use Stable Diffusion~\cite{rombach2022ldm} as the base model and adopt BrushNet~\cite{ju2024brushnet} and ControlNet~\cite{zhang2023controlnet} as the reference networks for editing.
We adopt the DDIM~\cite{song2020ddim} scheduler and perform denoising for 50 steps.
By default, the proposed correspondence-guided denoising strategy is applied from $4^{th}$ to $40^{th}$ steps and from the eighth attention layer to ensure consistency as well as preserve the strong generative prior.
Note the optimal choice of these may vary when different base models are utilized. 
The testing samples are partially acquired from the internet, while others of them are from the dataset of DreamBooth~\cite{ruiz2023dreambooth} and Custom Diffusion~\cite{kumari2023multi}.

\noindent\textbf{Evaluation metrics}.
We follow Custom Diffusion~\cite{kumari2023multi} and adopt the prevalent multi-modal model CLIP~\cite{radford2021clip} to evaluate various methods in terms of text alignment (TA) and editing consistency (EC).
Specifically, on the one hand, feature similarity of the target prompt and model output are computed to judge the textual alignment.
On the other hand, feature similarity of the edited images is adopted to evaluate the editing consistency.
User studies (US) are also incorporated to further evaluate the practical applicability and user satisfaction.

\noindent\textbf{Baselines.}
We include both local and global editing tasks, as well as numerous previous image editing methods, for comprehensive comparisons. Specifically,
for the task of local editing, we include prior works of Adobe Firefly~\cite{Firefly}, Anydoor~\cite{chen2024anydoor}, and Paint-by-Example~\cite{yang2023paint} for comparison.
Among these aforementioned methods, Firefly is a state-of-the-art commercial inpainting tool developed by Adobe, which could repaint local regions of the input image following the given textual prompts.
In order to achieve the task of consistent editing, the images among the set would be inpainted with the same detailed prompts.
Both Anydoor and Paint-by-example are Latent Diffusion Models (LDMs) supporting repainting target regions with the given reference image.
Thus, we sent an inpainted image to these models as the reference, expecting consistent editing results.
While for global editing, we compare our approach with MasaCtrl~\cite{cao2023masactrl}, StyleAlign~\cite{hertz2024stylealign}, and Cross-image attention~\cite{alaluf2024cross}.
The aforementioned methods achieve editing by manipulating and fusing attention features from various sources. 
Different from our method, they compute implicit correspondences from attention weights to ensure consistency among the editing outputs.

\subsection{Evaluation}
\label{subsec:evaluation}

\noindent\textbf{Qualitative results.}
We present a qualitative evaluation of the consistency editing methods, focusing on both local editing (image inpainting) and global editing (image translation). 
The comparisons for local editing in ~\cref{fig:local_comp} include results from our method, Adobe Firefly (AF), Anydoor (AD), and Paint-by-Example (PBE).
The results demonstrate that our approach consistently maintains the integrity of the input images across different modifications, including the cloth textures, mask and collar appearance, and even the eyelet amount of shoes, thanks to the introduction of explicit correspondence.
The baselines of global editing mainly include the ones predicted merely by implicit attentions - MasaCtrl (MC), StyleAligned (SA), and Cross-Image-Attention (CIA).
As in~\cref{fig:global_comp}, our method also achieves superior consistency and thematic adherence among the edits, such as the dress of the cat.
The implicit alternatives like MasaCtrl fail in the car roof, the high neckline of the elf, and the hole number of the robot.

\noindent\textbf{Quantitative results.}
\begin{table}[t]
\small
  \caption{Quantitative results respectively on local and global editing.
  We follow Custom Diffusion~\cite{kumari2023multi} to evaluate various methods on text alignment (TA) and editing consistency (EC).
  }
\centering
\begin{subtable}[b]{0.48\linewidth}
 \setlength{\tabcolsep}{5pt}
 \centering
\begin{tabular}{c c c c}
\toprule
Method & TA~$\uparrow$ & EC~$\uparrow$ \\
\midrule
AF~\cite{Firefly}  & 0.3082 & 0.8569 \\
AD~\cite{chen2024anydoor}  & 0.2981  & 0.8320  \\
PBE~\cite{yang2023paint} & 0.2969 & 0.8683 \\
Ours & \textbf{0.3176} & \textbf{0.8931} \\
\bottomrule
\end{tabular}
\end{subtable}\hfill
 \begin{subtable}[b]{0.48\linewidth}
 \setlength{\tabcolsep}{5pt}
 \centering
\begin{tabular}{c c c c}
\toprule
Method & TA~$\uparrow$ & EC~$\uparrow$\\
\midrule
MC~\cite{cao2023masactrl}    & 0.3140 & 0.9258  \\
SA~\cite{hertz2024stylealign}  & 0.3021  & 0.9099  \\
CIA~\cite{alaluf2024cross} & 0.2914 & 0.8912  \\
Ours & \textbf{0.3228} & \textbf{0.9355}\\
\bottomrule
\end{tabular}
\end{subtable}
  \label{tab:quantitative}
\end{table}
\begin{figure}[t]
    \centering
    \includegraphics[width=1.0\linewidth]{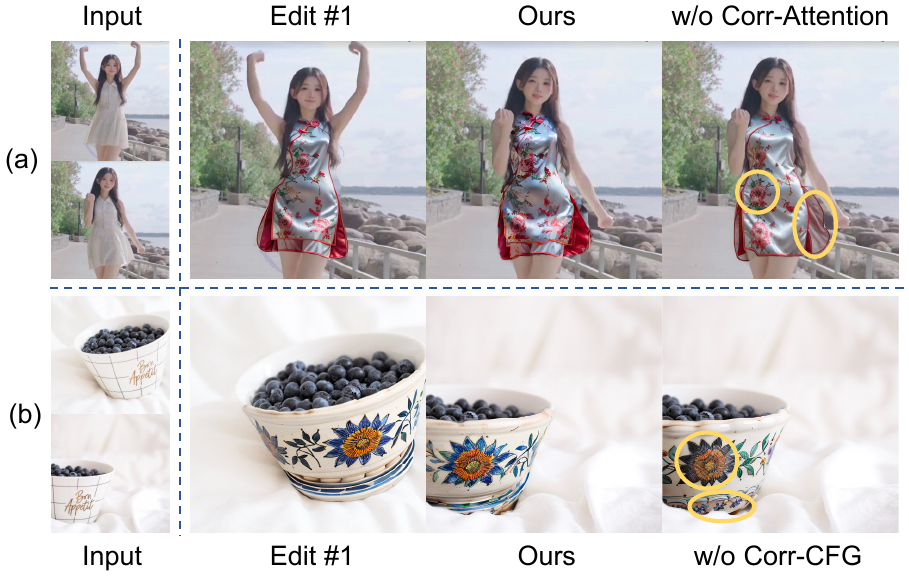}
    \caption{%
        Ablation studies on the (a) correspondence-guided attention manipulation (Corr-Attention) and (b) correspondence-guided CFG (Corr-CFG).
    }
    \label{fig:ablation}
    \vspace{-4pt}
\end{figure}
We conducted a comprehensive quantitative evaluation of our proposed method against several state-of-the-art image editing techniques, focusing on metrics of text alignment (TA) and editing consistency (EC) mentioned in~\cref{subsec:settings}. 
As illustrated in~\cref{tab:quantitative}, for local editing, our method attained the best scores in both TA and EC for local editing tasks, demonstrating a significant improvement over the competing methods.
For global editing tasks, our method continued to outperform the other counterparts, reaching a TA score of 0.3228 and an EC score of 0.9355. 
Overall, these results, along with the user studies in~\supp, clearly demonstrate the effectiveness of our method in achieving high text alignment and editing consistency across both local and global editing scenarios.

\subsection{Ablation Studies}
\label{subsec:ablation}

\begin{figure}[t]
    \centering
    \includegraphics[width=1.0\linewidth]{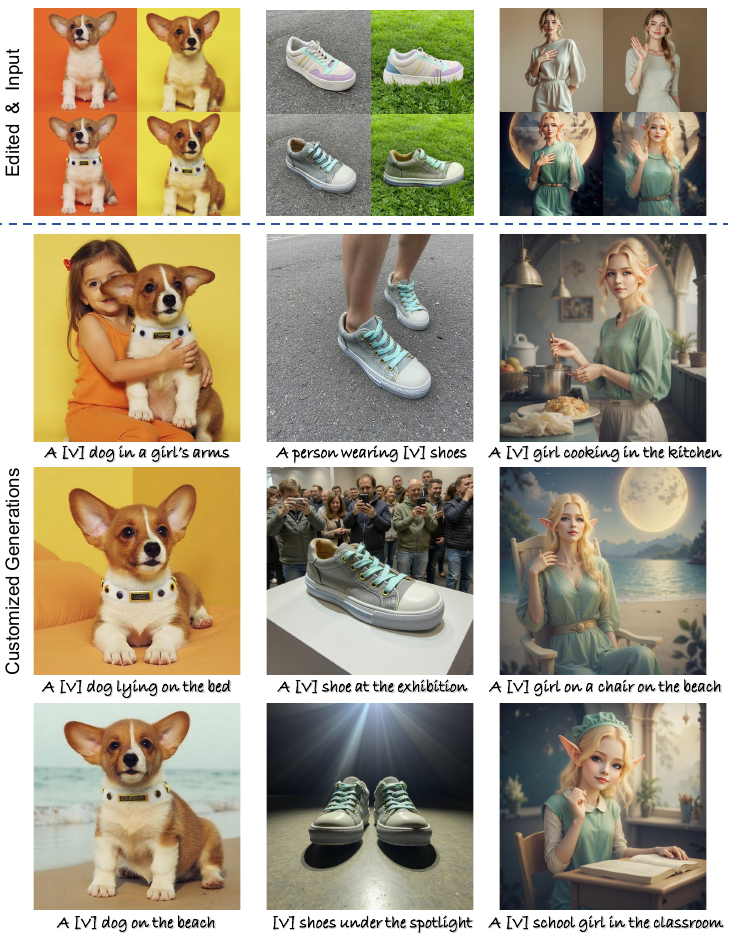}
    \caption{%
    With outputs from our consistent editing method (upper) and the customization~\cite{ruiz2023dreambooth} techniques,
    customized generation (lower) could be achieved by injecting the edited concepts into the generative model.
    }
    \label{fig:customization}
    \vspace{-6pt}
\end{figure}

\begin{figure*}[t]
    \centering
    \includegraphics[width=1.0\textwidth]{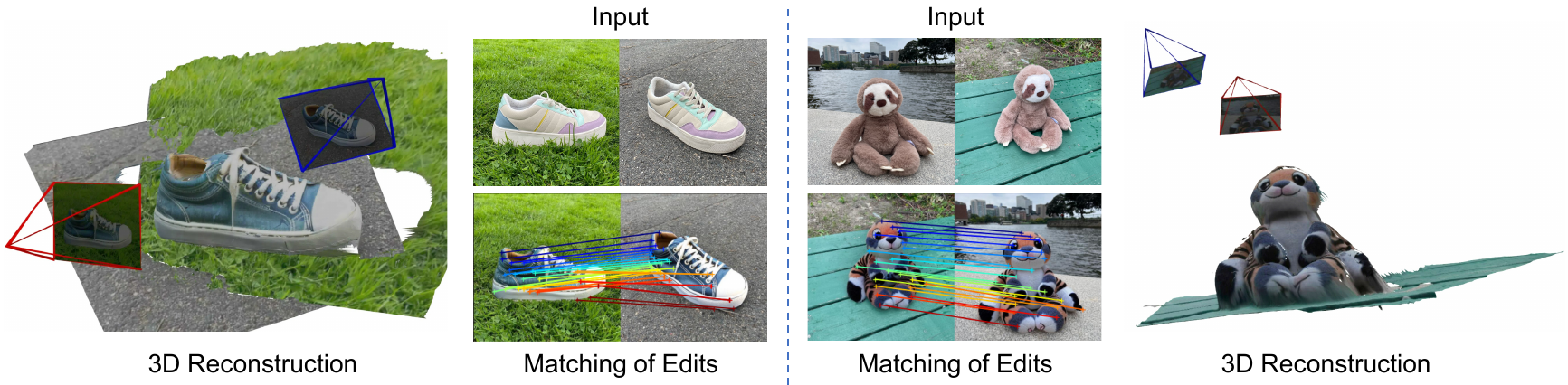}
    \caption{%
        We adopt the neural regressor Dust3R~\cite{wang2024dust3r} for 3D reconstruction based on the edits by matching the 2D points in a 3D space.
    }
    \label{fig:dust3r}
    \vspace{-5pt}
\end{figure*}

\begin{figure}[t]
    \centering
    \includegraphics[width=1.0\linewidth]{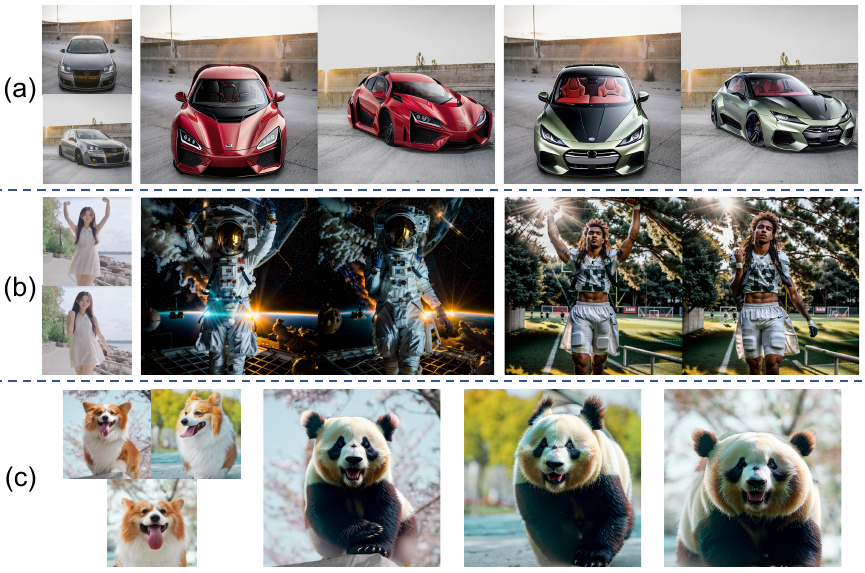}
    \caption{%
        Diverse results of consistent image inpainting (a) and translation (b) by the proposed method.
        Editing results for an image set of three images are demonstrated in (c).
    }
    \label{fig:diverse}
    \vspace{-8pt}
\end{figure}

In order to validate the effectiveness of the proposed correspondence-guided attention manipulation (Corr-Attention) and correspondence-guided CFG (Corr-CFG) introduced in~\ref{subsec:corr_attention} and~\ref{subsec:corr_CFG},
we conduct ablation studies by respectively disabling each of them and testing on the task of consistent editing.
When the proposed correspondence-guided attention manipulation (Corr-Attention) is disabled, the diffusion model relies on implicit attention correspondence to maintain consistency just like previous methods~\cite{cao2023masactrl, alaluf2024cross}.
As demonstrated in~\cref{fig:ablation} (a), the generative model then would yield flowers of wrong amounts and at improper locations.
The number of flowers and inconsistent textures speak for the effectiveness of introducing explicit correspondence to attention manipulation.
Recall that correspondence-guided CFG (Corr-CFG) is designed for finer consistency control functioning in the latent space of LDMs, which is validated in~\cref{fig:ablation} (b) where Corr-CFG achieves in generating more consistent textures for the flower on the bowl and stripes at the bottom of the bowl.
Additional ablation studies on attention manipulation and CFG could be found in~\supp.

\subsection{Additional Applications and Results}

\noindent\textbf{Customization based on the consistent edits.}
To further demonstrate the practical utility of the proposed method, we present an application example that integrates DreamBooth~\cite{ruiz2023dreambooth} and Low-Rank Adaptation (LoRA)~\cite{hu2021lora} techniques for customized image generation based on the multiple edited images.
Leveraging the edited outputs from our method, we employ DreamBooth to fine-tune a generative model for 500 steps for concept injection.
We also integrate LoRA techniques to further enhance the efficiency of this process by introducing low-rank matrices as adaptation parameters.
As in~\cref{fig:customization}, the fine-tuned generative model could yield desirable images corresponding to the edits after concept injection.
Thus novel concept generation and concept editing can be achieved in this way, serving as an application example of consistent editing.

\noindent\textbf{3D reconstruction based on the consistent edits.}
Furthermore, consistent editing could also benifit 3D reconstruction of the edits.
We achieve 3D reconstruction with a neural regressor~\cite{wang2024dust3r} which could predict accurate 3D
scene representations from the consistent image pairs.
Taking the edited images as inputs, the learned neural regressor could predict the 3D point-based models and 2D matchings without additional inputs such as camera parameters.
The reconstruction and matching results are presented in~\cref{fig:dust3r}, both of which also suggest the editing consistency of the proposed method.
The regressor respectively obtained 11,515 and 13,800 pairs of matching points for the two groups of edits, and only a portion are visualized for clear understanding.

\noindent\textbf{Additional results.}
Diverse results of multi-image inpainting and translation by the proposed method are provided in~\cref{fig:diverse} (a) and (b).
Editing results for an image set including three images are demonstrated in~\cref{fig:diverse} (c).
\section{Conclusion}
We introduce \method, a novel training-free method for consistent image editing across various images by leveraging explicit correspondence between them. 
Our approach enhances the self-attention mechanism and the classifier-free guidance computation by integrating correspondence information into the denoising process to ensure consistency among the edits.
The plug-and-play nature of our method allows for seamless integration into various models and its applicability across a wide range of tasks.
For limitations, sometimes the generated textures would be inconsistent due to the correspondence misalignment, which could be expected to be improved with better correspondence extractors.
And inheriting from the pre-trained editing models, sometimes distorted textures would be generated.
\clearpage
\newpage
{
\small
\bibliographystyle{ieeenat_fullname}
\bibliography{ref.bib}
}

\clearpage
\appendix
\renewcommand\thesection{\Alph{section}}
\renewcommand\thefigure{S\arabic{figure}}
\renewcommand\thetable{S\arabic{table}}
\renewcommand\theequation{S\arabic{equation}}
\setcounter{figure}{0}
\setcounter{table}{0}
\setcounter{equation}{0}
\setcounter{page}{1}
\maketitleappendix

\section*{Appendix}

\section{Overview}\label{sec:supp_overview}

We first illustrate the additional implementation details in~\cref{sec:supp_details}.
Additional ablation studies are also included in~\cref{sec:supp_ablation} to support the effectiveness of the designed components.
We also provide additional correspondence prediction comparisons in~\cref{sec:supp_corr}.
While~\cref{sec:supp_user_study} includes the user studies to validate the effectiveness of the proposed method in terms of user preference.
\cref{sec:supp_results} presents the additional qualitative results of the proposed method.

\section{Implementation Details}\label{sec:supp_details}

We adopt DDIM~\cite{song2020ddim} and perform denoising for 50 steps at the resolution of 512.
The proposed correspondence-guided denoising strategy is applied from $4^{th}$ to $40^{th}$ steps of the denoising process and from the eighth attention layer.
$\lambda$ and $\gamma$ are respectively set to be 0.8 and 0.9.
BrushNet~\cite{ju2024brushnet} and ControlNet~\cite{zhang2023controlnet} are adopted as the reference network for the task of consistent local and global editing.
We conduct on our experiments on a single A6000 GPU.

\section{Additional Ablation Studies}\label{sec:supp_ablation}
\vspace{-7pt}
\begin{figure}[h]
    \centering
    \includegraphics[width=1.0\linewidth]{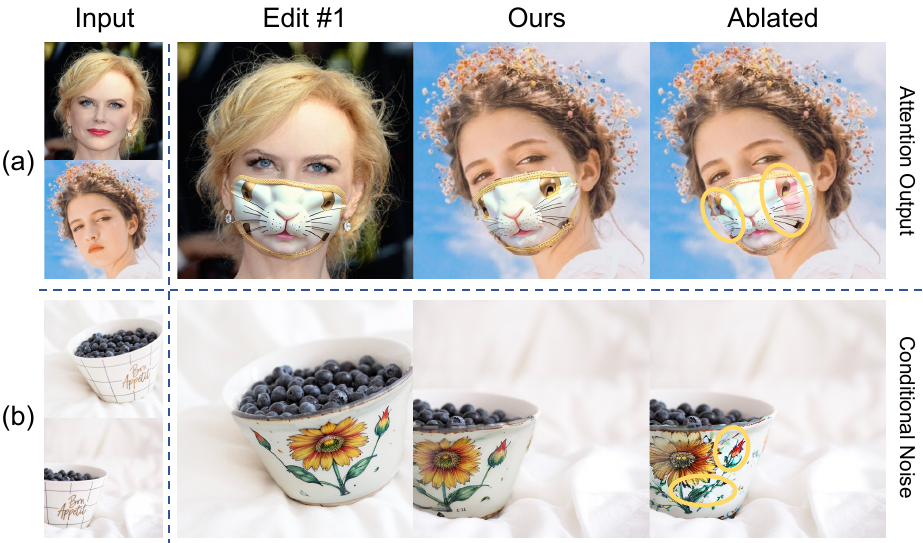}
    \caption{%
    Additional ablations on the correspondence-guided attention (upper) and CFG (lower).
    }
    \label{fig:supp_ablation}
\end{figure}
To further validate the effectiveness of the proposed components, we conduct experiments to ablate the correspondence-guided attention (upper) and CFG (lower).
As in~\cref{fig:supp_ablation} (a), we first modify the correspondence-guided attention manipulation to warp the attention outputs instead of the queries.
The distorted and inconsistent textures demonstrate that warping attention queries could better preserve the generative prior and achieve consistent results of high quality.
\cref{fig:supp_ablation} (b) demonstrates ablation results on the correspondence-guided CFG, where we guide the CFG in both conditional and unconditional noisy latents (instead of the unconditional only).
The generation result of the ablated version turns out to be unnatural and has a fragmented look with chaotic textures, suggesting the superiority of our design in correspondence-guided CFG, which avoids deteriorating the prior by merely manipulating the unconditional latents.

\section{Additional Correspondence Comparisons}\label{sec:supp_corr}
\vspace{-7pt}
\begin{figure}[h]
    \centering
    \includegraphics[width=1.0\linewidth]{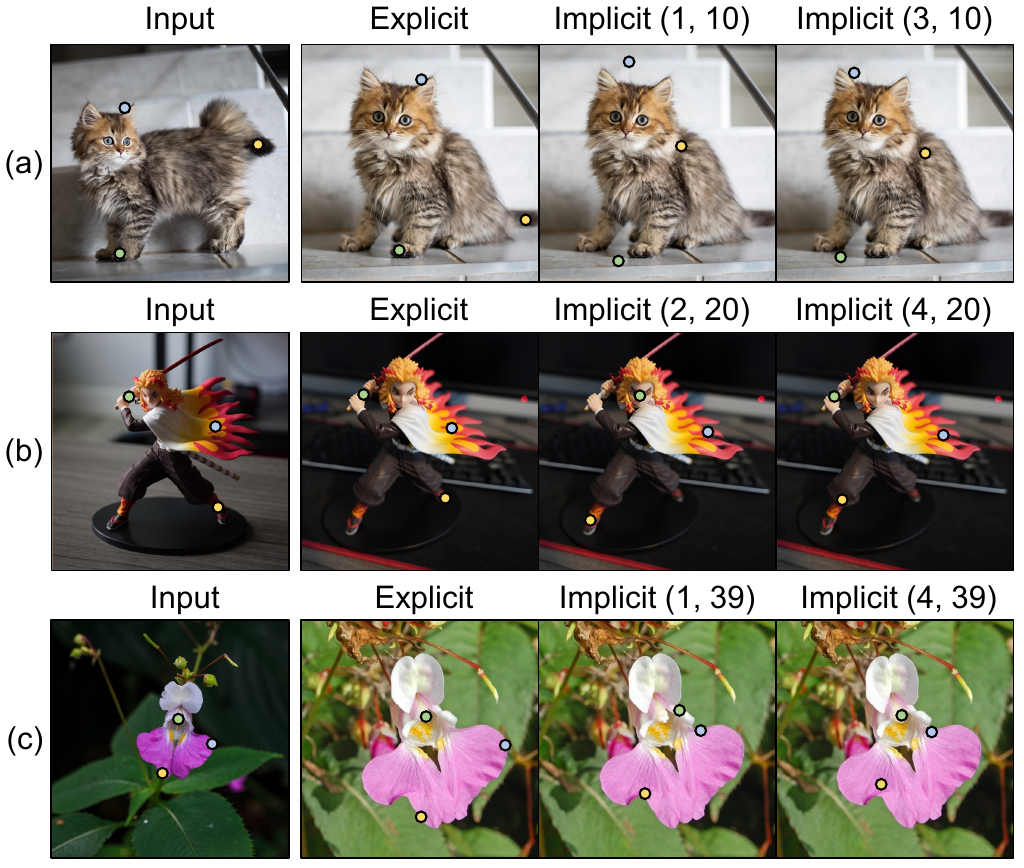}
    \caption{%
    Additional correspondence prediction comparisons.
    The numbers behind ``Implicit" respectively indicate the network layer and denoising step for correspondence prediction.
    }
    \label{fig:supp_corr1}
\end{figure}
We also incorporate additional comparing results of explicit and implicit correspondence prediction.
As in the main submission, the explicit ones are predicted with DIFT~\cite{tang2023dift} and implicit ones are computed by querying the attention keys of the target images with the source attention queries following~\cite{alaluf2024cross, cao2023masactrl}.
Specifically, we first perform image inversion~\cite{song2020ddim} and extract queries and keys of input images from the decoder of Stable Diffusion~\cite{rombach2022ldm}, as in~\cite{alaluf2024cross}.
Then the attention features are upsampled and the similarity is calculated based on the attention operation mentioned in the main submission.
During this experiment, we change the network layer and denoising step attention feature extraction for comprehensive studies.
These ranges are selected considering~\cite{alaluf2024cross} performs editing during steps of 10-39 and decoder layers of 1-4.
As in~\cref{fig:supp_corr1}, the implicit correspondence turns out to be less accurate, degrading the consistent editing, where the numbers (x, y) behind ``Implicit" indicate the layer and denoising step.
\begin{figure}[t]
    \centering
    \includegraphics[width=1.0\linewidth]{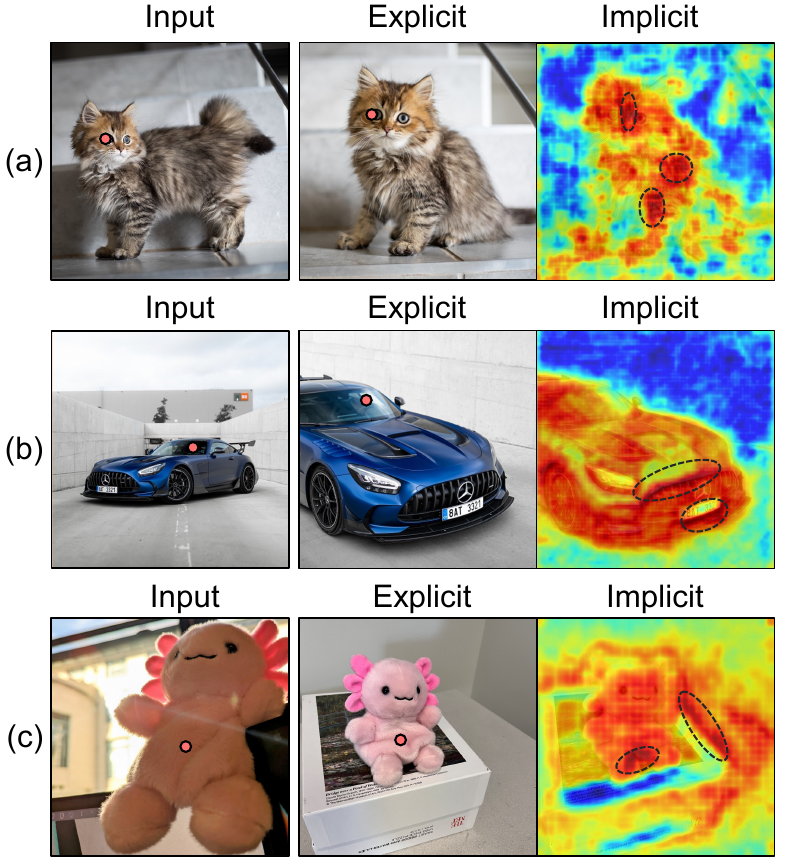}
    \caption{
    Additional correspondence prediction results with attention visualization. Regions with the highest attention weights are outlined with dashed circles.
    }
    \label{fig:supp_corr2}
\end{figure}

We further follow StyleAligned~\cite{hertz2024stylealign} to visualize the attention map for a better understanding of the editing process of implicit methods.
We first select a point on the source image, and then compute the attention map based on the aforementioned attention features.
The regions with the highest attention weights are
outlined with dashed circles in~\cref{fig:supp_corr2}, which suggests the implicit methods would query unreasonable regions that would cause undesirable and inconsistent textures.
The attention features of (a), (b), and (c) are respectively extracted from the denoising step of 10, 20, and 35, where the total step is 50.

\section{User Studies}\label{sec:supp_user_study}
\begin{figure}[h]
    \centering
    \includegraphics[width=1.0\linewidth]{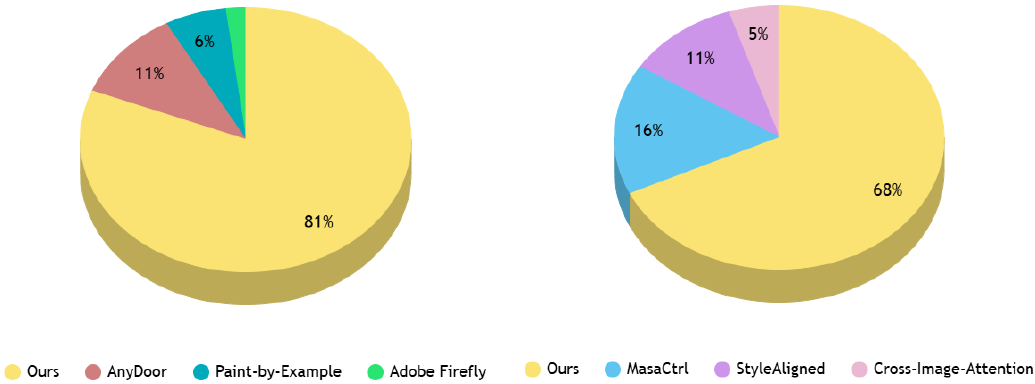}
    \caption{%
    User study results of consistent local editing (left) and global editing (right).
    }
    \label{fig:supp_user_study}
\end{figure}
As mentioned in the main submission, we conduct user studies to obtain results for user preference.
For the task of local and global editing, respectively, the 30 individuals are asked to finish up to 20 questions where they choose the best option among the four provided based on overall consistency, generation quality, and instruction-following, resulting in 500 votes for each task.  
As in~\cref{fig:supp_user_study},  our proposed solution has garnered significantly more preference compared to existing alternatives. In both evaluated tasks, over 60\% of the participants opted for our approach. This endorsement from the user validates the practical value of our method as well as highlights its potential impact in real-world applications.

\section{Additional Results}\label{sec:supp_results}
\begin{figure*}[h]
    \centering
    \includegraphics[width=1.0\textwidth]{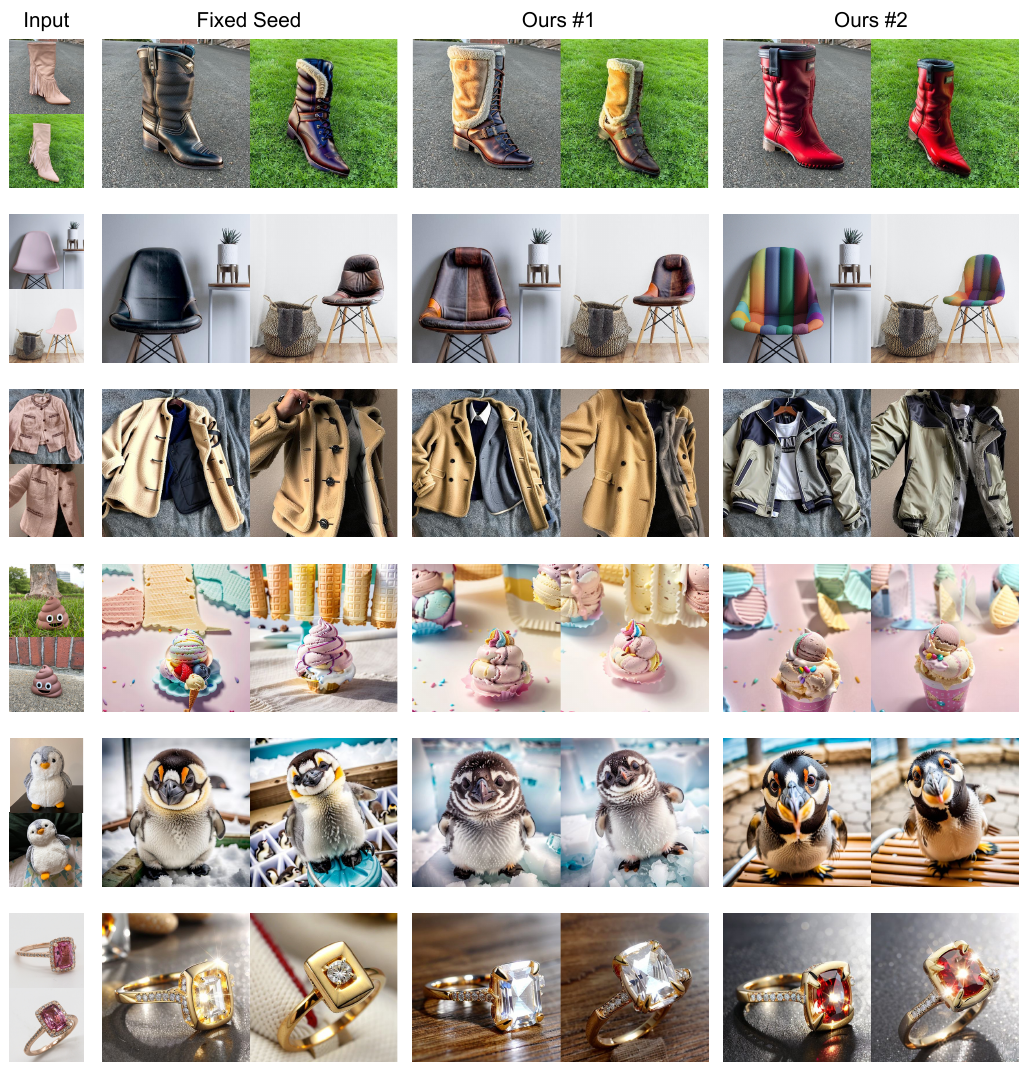}
    \caption{
    Additional qualitative results of the proposed method for local (upper three) and global editing (lower three ones).
    The inpainted regions for local editing are indicated with the
    light red color.
    ``Fixed Seed" indicates editing results from the same random seed (the same initial noise).
    }
    \label{fig:supp_qualitative}
\end{figure*}

For better understanding, we also incorporate additional qualitative results for consistent local and global editing in~\cref{fig:supp_qualitative}.
Editing results from the same initial noise are also provided in the figure, indicated as ``Fixed Seed".
The inpainted regions for local editing are indicated with light red color.

\end{document}